\documentclass{article}
\usepackage{amssymb,amsthm,amsmath}
\usepackage[ruled,vlined,linesnumbered]{algorithm2e}

\newtheorem{theorem}{Theorem}
\newtheorem{remark}{Remark}

\newcommand{\Vertt}{,}
\newcommand{\Ber}{\mathrm{Ber}}
\newcommand{\arm}[1]{i(#1)}
\newcommand*{\regret}{\mathrm{regret}}
\newcommand{\rd}{\mathrm{d}}
\newcommand{\per}{\,.}
\newcommand{\com}{\,,}
\newcommand{\since}[1]{\quad\left(\mbox{#1}\right)}
\newcommand{\nn}{\nonumber\\}
\newcommand{\n}{\nonumber}
\DeclareMathOperator*{\argmax}{\mathrm{argmax}}

\newcommand{\E}{\mathbb{E}}
\newcommand{\Ex}[1]{\mathbb{E}\left[#1\right]}
\newcommand{\id}{ \mbox{\rm 1}\hspace{-0.63em}\mbox{\rm \small 1\,}}
\newcommand{\idx}[1]{\id\left[#1\right]}
\newcommand{\muhat}{\hat{\mu}}
\newcommand{\ucb}{\overline{\mu}}
\newcommand{\pax}[1]{\left(#1\right)}
\newcommand{\sqx}[1]{\left[#1\right]}
\newcommand{\brx}[1]{\left\{#1\right\}}
\newcommand{\ep}{\epsilon}
\newcommand*{\e}{\mathrm{e}}
\newcommand{\lo}{\mathrm{O}}

\newcommand{\flooring}[1]{\left\lfloor#1\right\rfloor}

\title{A Note on KL-UCB+ Policy for the Stochastic Bandit}
\author{Junya Honda\thanks{Email: honda@stat.t.u-tokyo.ac.jp}\\
The University of Tokyo / RIKEN}
\date{}

\begin{document}

\maketitle
\allowdisplaybreaks[4]

\begin{abstract}%
A classic setting of the stochastic $K$-armed bandit problem
is considered in this note.
In this problem it has been known that KL-UCB policy achieves the asymptotically optimal regret
bound and KL-UCB+ policy empirically performs better than the KL-UCB policy although
the regret bound for the original form of the KL-UCB+ policy has been unknown.
This note demonstrates that
a simple proof of the asymptotic optimality of the KL-UCB+ policy
can be given by the same technique as those used for analyses of other known policies.
\end{abstract}

\section{Overview}
In the problem of the stochastic bandit problems,
it is known that there exists a (problem-dependent) regret lower bound
\cite{lai}\cite{burnetas}.
It can be achieved by, for example, the DMED policy \cite{honda_colt} for the model of
nonparametric distributions over $[0,1]$.
On the other hand, the policy proposed in \cite{burnetas},
which is often called\footnote{The original paper by Burnetas and Katehakis \cite{burnetas} did not explicitly give a
name for the policy but it is referred to as Inflated Sample Mean (ISM) policy
by their group \cite{KL_CHK}.}
KL-UCB \cite{KL_UCB_journal}, achieves
the asymptotic bound for some models such as one-parameter exponential families and the family of distributions over a finite support.
One of the conference version \cite{KL_UCB} of
\cite{KL_UCB_journal} also proposed KL-UCB+ policy,
which empirically performs better than KL-UCB but does not have a theoretical guarantee.

The KL-UCB+ policy is obtained by replacing $t$
in the confidence bound with $t/N_i(t)$, where
$t$ is the current number of trials and $N_i(t)$ is the number of samples from the arm $i$.
It is discussed in \cite{KL_UCB} that KL-UCB+ can be related to DMED, which also has term $t/N_i(t)$
in the criterion for choosing the arm to pull, and \cite{KL_UCB}
used the name DMED+ for the same policy based on this observation.

After these works, IMED policy is proposed \cite{honda_imed} as an index-policy version of the DMED in \cite{honda_colt} (or equivalently, DMED+ in \cite{KL_UCB_journal}),
which achieves the asymptotic bound and empirically performs almost the same as the KL-UCB+ policy
with low computational cost.
On the other hand, the asymptotic optimality of the KL-UCB+ has not been proved
to the best of the author's knowledge,
although proofs are given for its variants \cite{KLUCBp_kaufmann}\cite{KLUCBpp}.

In this note, we show that
this difference,
known optimality of DMED/IMED and unknown optimality of KL-UCB+,
simply comes from the difference of the used techniques
between \cite{honda_colt}\cite{honda_imed} and \cite{KL_UCB_journal} by
demonstrating that the asymptotic optimality of (a slightly generalized version of) KL-UCB+ can be proved by exactly the same argument as those for DMED and IMED.
To be more specific, whereas the typical technique for KL-UCB such as \cite{KL_UCB_journal}\cite{KL_maillard}
reduces the regret analysis to the evaluation of a hitting probability
of a boundary for some stochastic process,
the technique in \cite{honda_colt}\cite{honda_imed}
reduces it to the evaluation of the expected waiting-time
for the optimal arm to be pulled the next time.
These two analyses incur different looseness arising from double-counting
of events and there is no clear winner,
but the latter one seems to be convenient for the proof of the asymptotic optimality of KL-UCB+.

Note that the original analyses in \cite{honda_colt} and \cite{honda_imed}
are given for general distributions over $[0,1]$ and $(-\infty,1]$, respectively, but
this note only considers Bernoulli distributions to simply illustrate
the difference of the regret decomposition between \cite{honda_colt}\cite{honda_imed} and \cite{KL_UCB_journal}.

It should be noted that this result does not mean that for any model KL-UCB+ always achieves the asymptotic bound only if KL-UCB does.
In fact, it seems to be almost hopeless to expect the asymptotic optimality of KL-UCB+
for general multi-parameter models with a non-compact parameter space from the discussions in \cite{KL_CHK} and \cite{KL_maillard}.

\section{Problem Setting}
Let there be $K\in\mathbb{N}$ arms.
The agent pulls one arm $\arm{t}\in\{1,2,\cdots,K\}$ and
observe reward $r(t)\in\{0,1\}$
at each round $t=1,2,\dots$.
The observed reward from the arm $i$ independently follows
Bernoulli distribution $\Ber(\mu_i)$, where the expected reward $\mu_i\in[0,1]$
is unknown to the agent.

The objective of the agent is to maximize the cumulative reward through $T\in\mathbb{N}$ rounds,
and its performance is measured by {\it regret} or {\it pseudo-regret},
which is given by
\begin{align}
\regret(T)=
\sum_{t=1}^{T}(\mu^*-\mu_{\arm{t}})
\com\n
\end{align}
where $\mu^*=\mu_{i^*}$ for $i^*=\argmax_{i\in\{1,2,\cdots,K\}}\mu_i$.
It is shown in \cite{lai} that any policy satisfying a mild regularity condition
called concistency has regret lower bound
\begin{align}
\liminf_{T\to\infty}\frac{\E[\regret(T)]}{\log T}\ge
\sum_{i\neq i^*}\frac{\mu^*-\mu_i}{d(\mu_i\Vertt \mu^*)}
\com\n
\end{align}
where $d(x\Vertt y)=x\log (x/y)+(1-x)\log ((1-x)/(1-y))$ is the KL divergence between Bernoulli distributions
$\Ber(x)$ and $\Ber(y)$.

Let $N_i(t)$ denote the number of rounds that the arm $i$ is pulled
before the $t$-th round.
We define $\muhat_i(t)=(1/N_i(t))\sum_{t'=1}^{t-1}\idx{\arm{t}=i}r(t)$ as the empirical mean of the rewards from arm $i$ before the $t$-th round
and $\muhat_{i,n}$ as the empirical mean of the rewards from the arm $i$ when
the arm $i$ is pulled $n$ times.
Then we have $\muhat_{i,N_i(t)}=\muhat_{i}(t)$.

We consider a slightly generalized version of KL-UCB+ policy with parameter $\alpha\ge 0$,
which we denote by KL-UCB$(\alpha)$ policy.
In this policy, the pulled arm is the one maximizing the UCB score given by
\begin{align}
\ucb_i(t)=\sup\brx{\mu\in[0,1]: d(\muhat_{i}(t)\Vertt \mu)\le \frac{\log \frac{t}{(N_{i}(t))^{\alpha}}}{N_{i}(t)}}\per\n
\end{align}
More formally, the KL-UCB$(\alpha)$ policy is given by Algorithm~\ref{klucb}.
\begin{algorithm}[t]
\DontPrintSemicolon
\SetKwInput{Input}{Input}
\SetKwInput{Parameter}{Parameter}
\Parameter{$\alpha\ge 0$.}
Pull each arm once.\;
\For{$t=K+1,K+2,\cdots,T$}{
 Pull arm $\arm{t}=\argmax_{i\in[K]}\ucb_i(t)$.
}
\caption{KL-UCB$(\alpha)$ Policy}\label{klucb}
\end{algorithm}%
Here $\alpha=0$ corresponds to the KL-UCB policy
and $\alpha=1$ corresponds to the KL-UCB+ policy.
Here note that \cite{KLUCBp_kaufmann} proved the asymptotic optimality
for the modified version with an extra exploration term such that $t/(N_i(t))^{\alpha}$ is replaced
with $(t\log^c t)/N_i(t)$ with $c\ge 7$.

\section{Result}
In this note we prove the following regret bound.
\begin{theorem}
Assume $\mu^*<1$.
The expected regret of the KL-UCB$(\alpha)$ policy
is bounded for any $\ep\in(0,(\mu^*-\max_{i\neq i^*}\mu_i)/2)$
by
\begin{align}
\E[\regret(T)]
&\le
\sum_{i\neq i^*}
(\mu^*-\mu_i)\pax{n_i+\frac{1}{2\ep^2}}
+
\e\Gamma(\alpha+2)\pax{1+\log \frac{1}{1-\mu^*}}
\pax{\frac{\mu^*(1-\mu^*+\ep)}{\ep^2}}^{\alpha+2}\com\label{bound1}
\end{align}
where $\Gamma(\cdot)$ is the gamma function and
$n_i=\sup\{x\ge 0: x^{\alpha}\e^{xd(\mu_i+\ep\Vertt \mu^*-\epsilon)}\le T\}$.
\end{theorem}
\begin{remark}{\rm
The proof only uses the fact that the reward is in $[0,1]$
as a property of Bernoulli distributions.
Thus the same regret bound holds for general distributions on $[0,1]$
like the kl-UCB policy.
}\end{remark}
\begin{remark}{\rm
The first term $n_i$ in \eqref{bound1}
is trivially bounded by
\begin{align}
n_i\le \frac{\log T}{d(\mu_i+\ep\Vertt \mu^*-\ep)}\com\n
\end{align}
which leads to the asymptotic bound
\begin{align}
\limsup_{T\to\infty}\frac{\E[\regret(T)]}{\log T}
&\le
\sum_{i\neq i^*}
\frac{\mu^*-\mu_i}{d(\mu_i\Vertt \mu^*)}
\per\n
\end{align}
We can also express $n_i$ as
\begin{align}
n_i
&=
\frac{\alpha}{d(\mu_i+\ep\Vertt \mu^*-\ep)}
W_0\pax{\frac{T^{\frac{1}{\alpha}}d(\mu_i+\ep\Vertt \mu^*-\ep)}{\alpha}}\com\n
\end{align}
where $W_0(x)$ is Lambert $W$ function, that is, the solution of $x=z \e^z,\,z\ge -1$.
The expansion $W_0(x)=\log x-\log \log x+\lo\pax{\frac{\log \log x}{\log x}}$
\cite[Sect.~4.13]{nist_handbook} for large $x$ leads to the bound
\begin{align}
n_i=
\frac{\log T-\alpha \log \log T}{d(\mu_i+\ep\Vertt \mu^*-\ep)}
+\lo\pax{\frac{\log \log T}{\log T}}\per\label{bound2}
\end{align}
Note that the derivation of \eqref{bound1} is essentially the same as \cite[Theorem 3]{honda_imed}
for the IMED policy, which suffers from heavy dependence on $\epsilon$.
On the other hand, \cite[Theorem 5]{honda_imed} gives a more complicated but tighter bound for this policy.
We can expect that the same technique combined with \eqref{bound2} can be used to obtain a higher-order bound
\begin{align}
\limsup_{T\to\infty}\frac{\E[\regret(T)]-\sum_{i\neq i^*}
\frac{(\mu^*-\mu_i)\log T}{d(\mu_i\Vertt \mu^*)}
}{\log \log T}
&<0\n
\end{align}
for the KL-UCB$(\alpha)$ policy with $\alpha>0$, although it makes the analysis much longer.
}\end{remark}

\begin{proof}
First we bound the regret as
\begin{align}
\regret(T)
&=
\sum_{t=1}^T \pax{\mu^*-\mu_{\arm{t}}}\nn
&\le
\sum_{i\neq i^*}(\mu^*-\mu_i)
\underbrace{\sum_{t=1}^T
\idx{\arm{t}=i,\,\ucb^*(t)\ge \mu^*-\ep}
}_{\mathrm{(A)}}
+
\underbrace{\sum_{t=1}^T
\idx{\ucb^*(t)< \mu^*-\ep}
}_{\mathrm{(B)}}\com\label{ucb_bunkai}
\end{align}
where we write
$\ucb^*(t)=\max_{i\in\{1,2,\cdots,K\}}\ucb_i(t)$.

The term (A) is expressed as
\begin{align}
\mathrm{(A)}
&=
\sum_{n=1}^T\sum_{t=1}^T \idx{\arm{t}=i,\,\ucb^*(t)\ge \mu^*-\epsilon,N_i(t)=n}\nn
&=
\sum_{n=1}^T \idx{\bigcup_{t=1}^T\left\{\arm{t}=i,\,\ucb^*(t)\ge \mu^*-\epsilon,N_i(t)=n\right\}}\com\n
\end{align}
where we used the fact that the event
$\{N_i(t)=n,\,\arm{t}=i\}$ occurs at most once for each $(n,i)$.
Since
$\arm{t}=i$ implies $\ucb_i(t)=\ucb^*(t)$,
we have
\begin{align}
\mathrm{(A)}
&\le
\sum_{n=1}^T \idx{\bigcup_{t=1}^T\left\{\ucb_i(t)=\ucb^*(t)\ge\mu^*-\epsilon,N_i(t)=n\right\}}\nn
&\le
\sum_{n=1}^T \idx{\bigcup_{t=1}^T\left\{
\sup\brx{\mu: d(\muhat_i(t)\Vertt \mu)\le \frac{\log \frac{t}{(N_i(t))^{\alpha}}}{N_i(t)}}
\ge \mu^*-\epsilon,N_i(t)=n\right\}}\nn
&\le
\sum_{n=1}^T \idx{
\sup\brx{\mu: d(\muhat_{i,n}\Vertt \mu)\le \frac{\log \frac{T}{n^{\alpha}}}{n}}
\ge \mu^*-\epsilon}\nn
&=
\sum_{n=1}^T \idx{
\brx{n^{\alpha}\e^{nd(\muhat_{i,n}\Vertt \mu^*-\epsilon)}\le T }\cup
\brx{\muhat_{i,n}\ge \mu^*-\ep}
}\nn
&\le
n_i
+
\sum_{n=\flooring{n_i}+1}^T
\idx{
\brx{d(\muhat_{i,n}\Vertt \mu^*-\epsilon)\le d(\mu_i+\ep\Vertt \mu^*-\ep)}
\cup
\brx{\muhat_{i,n}\ge \mu^*-\ep}
}\nn
&\qquad\qquad\qquad\since{by definition $n_i=\sup\{x\ge 0: x^{\alpha}\e^{xd(\mu_i+\ep\Vertt \mu^*-\epsilon)}\le T\}$}\nn
&\le
n_i
+
\sum_{n=\flooring{n_i}+1}^T
\idx{
\muhat_{i,n}\ge \mu_i+\ep
}
\per\n
\end{align}
Therefore, by Hoeffding's inequality we have
\begin{align}
\E[\mathrm{(A)}]
&\le
n_i
+
\sum_{n=1}^{\infty}\Pr[\muhat_{i,n}\ge \mu_i+\ep]\nn
&\le
n_i
+
\sum_{n=1}^{\infty}\e^{-2n\ep^2}\nn
&=
n_i
+
\frac{1}{\e^{2\ep^2}-1}
\le
n_i
+
\frac{1}{2\ep^2}\com
\label{ucb_hoeffding}
\end{align}
where we used
the fact $\e^{x}\ge 1+x$.

Next we evaluate the term (B).
By the definition of the UCB score we have
\begin{align}
\{\ucb_{i^*}(t)< \mu^*-\epsilon\}
&\Leftrightarrow
\left\{
\sup\brx{\mu: d(\muhat_{i^*}(t)\Vertt \mu)\le \frac{\log \frac{t}{(N_{i^*}(t))^{\alpha}}}{N_{i^*}(t)}}
<\mu^*-\ep\right\}\nn
&\Rightarrow
\left\{
d(\muhat_{i^*}(t)\Vertt \mu^*-\ep)> \frac{\log \frac{t}{(N_{i^*}(t))^{\alpha}}}{N_{i^*}(t)}
,\,
\muhat_{i^*}(t)<\mu_{i^*}-\epsilon
\right\}\nn
&\Leftrightarrow
\left\{
t< (N_{i^*(t)})^{\alpha}\e^{N_{i^*}(t)d(\muhat_{i^*}(t)\Vertt \mu^*-\ep)},\,
\muhat_{i^*}(t)<\mu_{i^*}-\epsilon
\right\}\per\n
\end{align}
Therefore,
\begin{align}
\mathrm{(B)}
&\le
\sum_{t=1}^T \idx{\ucb_{i^*}(t)< \mu^*-\epsilon}\nn
&\le
\sum_{n=1}^T\sum_{t=1}^T \idx{\ucb_{i^*}(t)<\mu^*-\epsilon,N_{i^*}(t)=n}\nn
&\le
\sum_{n=1}^T\sum_{t=1}^T \idx{t< n^{\alpha}\e^{nd(\muhat_{i^*}(t)\Vertt \mu^*-\ep)},\,
\muhat_{i^*,n}<\mu^*-\epsilon}\nn
&\le
\sum_{n=1}^T n^{\alpha}\e^{nd(\muhat_{i^*,n}\Vertt \mu^*-\ep)}\idx{\muhat_{i^*,n}< \mu^*-\epsilon}\label{essential}\\
&\le
\sum_{n=1}^T n^{\alpha}\e^{n\pax{d(\muhat_{i^*,n}\Vertt \mu^*)-\frac{\ep^2}{2\mu^*(1-\mu^*+\ep)}}}\idx{\muhat_{i^*,n}< \mu^*-\epsilon}\com\n
\end{align}
where the last inequality follows from
\begin{align}
d(x\Vertt \mu')-d(x\Vertt \mu)\ge \frac{(x-\mu)^2}{2\mu'(1-\mu)}
\label{yuru}
\end{align}
for $x\le \mu\le \mu'<1$
\cite[Lemma 13]{honda_colt}.
Here, for
$P_n(x)=\Pr[d(\muhat_{i^*,n}\Vertt \mu^*)\ge x,\,\muhat_{i^*,n}< \mu^*-\epsilon]$,
we have
$P_n(x)\le \e^{-x}$
by Chernoff-Hoeffding's inequality.
Letting
$\ep'=\ep^2/(2\mu^*(1-\mu^*+\ep))$ and
$d_1=d(0\Vertt \mu^*)=\log 1/(1-\mu^*)$, we obtain
\begin{align}
\E[(\mathrm{B})]
&\le
\sum_{n=1}^T\Ex{n^{\alpha}\e^{n\pax{d(\muhat_{i^*,n}\Vertt \mu^*)-\ep'}}\idx{\muhat_{i^*,n}\le \mu^*-\epsilon}}\nn
&=
\sum_{n=1}^T\int_{0}^{d_1}n^{\alpha}\e^{n\pax{x-\ep'}}
\rd (-P(x))\nn
&=
\sum_{n=1}^T
\pax{
\sqx{
-n^{\alpha}\e^{n\pax{x-\ep'}}
P(x)
}_{0}^{d_1}
+n^{\alpha+1}\int_{0}^{d_1}\e^{n\pax{x-\ep'}}
P(x)\rd x
}
\since{integration by parts}\nn
&\le
\sum_{n=1}^T
\left(
n^{\alpha}\e^{-n\ep'}+
n^{\alpha+1}\int_{0}^{d_1}\e^{-n\ep'}\rd x
\right)\nn
&\le
(d_1+1)
\sum_{n=1}^T
n^{\alpha+1}
\e^{-n\ep'}
\label{finite_int}\\
&\le
(d_1+1)
\int_0^{\infty}
(u+1)^{\alpha+1}
\e^{-u\ep'}\rd u
\nn
&=
(d_1+1)
\e^{\ep'}
\int_{\ep'}^{\infty}
\pax{\frac{v}{\ep'}}^{\alpha+1}\e^v\frac{\rd v}{\ep'}
\since{by letting $v=\ep'(u+1)$}
\nn
&\le
\frac{\e(d_1+1)
\Gamma(\alpha+2)}{(\ep')^{\alpha+2}}
\since{by $\ep'\le 1$}.
\label{ucb_b}
\end{align}
We complete the proof
by combining \eqref{ucb_hoeffding} and \eqref{ucb_b} with \eqref{ucb_bunkai}.
\end{proof}

\begin{remark}{\rm
In this proof we used two properties specific to the Bernoulli distributions:
the existence of constant $C(\mu,\mu')>0$
such that $\inf_{x\le \mu}\{d(x,\mu')-d(x,\mu)\}\ge C(\mu,\mu')$
in \eqref{yuru},
and finiteness of $\sup_{x}d(x\Vertt \mu)$ in \eqref{finite_int}.
At least one of them does not hold for most models, which makes
the analysis difficult especially for
multi-parameter models with non-compact parameter spaces,
where we need more sophisticated evaluation for
the expectation of \eqref{essential} such as \cite{honda_imed}.
Note that a term similar to \eqref{essential} also appears
in the regret lower bound of some policies.
Therefore the boundedness of the expectation
of \eqref{essential} is quite essential for the policy
to have a regret bound.
In fact, for the case of normal distributions
with unknown means and variances,
we can show
by evaluating a term similar to \eqref{essential}
that the (original form of) KL-UCB policy \cite{KL_CHK}
and Thompson sampling with Jeffreys prior \cite{ts_gauss}
incur polynomial regret.
}\end{remark}

\section*{Acknowledgement}
The author thanks Mr.~Kohei Takagi for his survey on the analysis of KL-UCB+.
The author also thanks Professor Vincent Y.~F.~Tan for finding many typos in the first version.

\end{document}